\documentclass[10pt,twocolumn,letterpaper]{article}

\usepackage[pagenumbers]{cvpr} 

\usepackage{graphicx}
\usepackage{amsmath}
\usepackage{amssymb}
\usepackage{booktabs}
\usepackage{adjustbox}
\usepackage{tabularx}
\usepackage{float}
\usepackage{lipsum}
\usepackage{mathtools}
\usepackage{cuted}
\usepackage{pifont}
\newcommand{\cmark}{\ding{51}}%

\usepackage[pagebackref,breaklinks,colorlinks]{hyperref}
 
\renewcommand*\backref[1]{\ifx#1\relax \else (Cited on page #1) \fi}
\newcommand{\squeezeup}{\vspace{-2mm}}

\usepackage[capitalize]{cleveref}
\crefname{section}{Sec.}{Secs.}
\Crefname{section}{Section}{Sections}
\Crefname{table}{Table}{Tables}
\crefname{table}{Tab.}{Tabs.}

\def\cvprPaperID{10856} 
\def\confName{CVPR}
\def\confYear{2023}

\begin{document}

\title{ALADIN-NST: Self-supervised disentangled representation learning of artistic style through Neural Style Transfer}

\author{Dan Ruta \\
University of Surrey \\
\\
\and
Gemma Canet Tarrés \\
University of Surrey \\
\\
\and
Alexander Black \\
University of Surrey \\
\\
\and
Andrew Gilbert \\
University of Surrey \\
\\
\and
John Collomosse\\
University of Surrey, Adobe \\
 \\
}

\maketitle

\begin{abstract}
Representation learning aims to discover individual salient features of a domain in a compact and descriptive form that strongly identifies the unique characteristics of a given sample respective to its domain. 
Existing works in visual style representation literature have tried to disentangle style from content during training explicitly. A complete separation between these has yet to be fully achieved. 
Our paper aims to learn a representation of visual artistic style more strongly disentangled from the semantic content depicted in an image. 
We use Neural Style Transfer (NST) to measure and drive the learning signal and achieve state-of-the-art representation learning on explicitly disentangled metrics. 
We show that strongly addressing the disentanglement of style and content
leads to large gains in style-specific metrics,
encoding far less semantic information and achieving state-of-the-art accuracy in downstream multimodal applications.
\end{abstract}

\section{Introduction}
\label{sec:intro}
Artistic style refers to the unique visual appearance of how a subject is depicted in a work of art. Style is ever-evolving, and it is complex, if not impossible, to create an exhaustive ontology for. Therefore, capturing this subjective information in a model is an open and challenging area of research.
Even for humans, style can be challenging to pinpoint and separate. However, this task is easier in a comparative setting, where similarities and differences between two stylistically similar images can hint at common properties. A constant challenge with automated approaches and human judgment is separating and disentangling style from the subject matter. This is especially an issue in the comparative case, where two stylistically similar images can often represent the same content.   

Yet a representation of style has many applications. Aside from simple style-based image retrieval tasks, there are also other uses, such as style conditioned image generation \cite{cogs,parasol}, stylization \cite{hypernst}, automatic style tagging/captioning \cite{stylebabel}, and image translation \cite{sae}.

Disentanglement in embeddings is critical in such multimodal applications, where a clean disentangled signal of the given modality is especially needed for aligning with other modalities. Thus, improvements to the disentanglement of embeddings for a modality such as artistic style can more cleanly expose only the desired style features without semantic information.

In our work,
we show that this style/content entanglement is still present in state-of-the-art representations.
We propose a novel learning algorithm for fully disentangled learning of style. This explicitly disentangled representation benefits downstream tasks like style-based image retrieval and tagging. Our contributions are \footnote{Our codebase: \url{https://github.com/DanRuta/aladin-nst}}: 

\begin{enumerate}
    \item A novel methodology for training a style representation model without any content/style entanglement in the data, trained over BBST-4M \cite{neat}
    \item New state-of-the-art in style representation learning with enforced disentanglement, with a new benchmark dataset 
    \item New state-of-the-art multimodal vision/language learning in the context of artistic style for automatic style tagging
\end{enumerate}

\begin{figure*}[h]
    \label{fig:data}
    \begin{center}
    \includegraphics[width=1\linewidth]{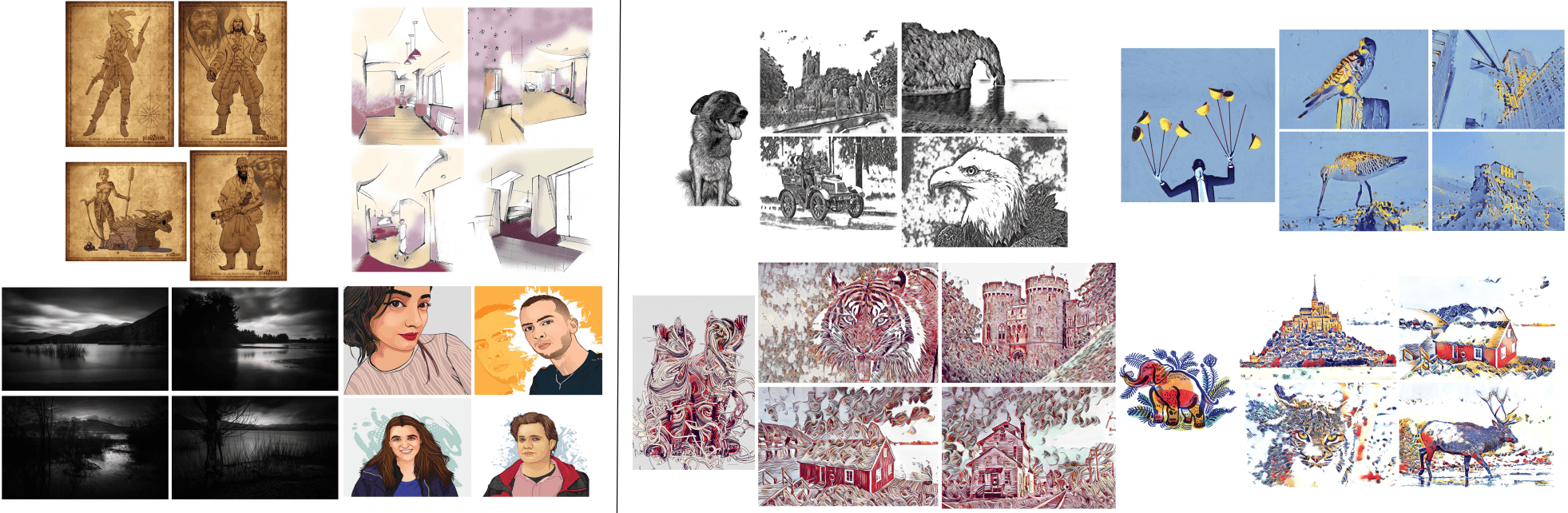}  
    
    \caption{
    Please zoom in for details.
    (Left) Example style groups from the BAM-FG dataset. The images in each group are style consistent, but they are also semantically consistent. For example, the top left style group has a consistent \textit{weathered paper} style but is also consistent in the subject matter of character design. The top right has consistent \textit{pastel} style but is consistently interiors. The bottom left is consistent \textit{moody vignette dark photography} style, but all images are of landscapes. Bottom right \textit{vector art} images all contain faces.
    (Right) Example synthetic style consistent images, as used in our work (via NeAT). The left-most images in each style group are the reference style image.
    The BAM-FG data (left) shows style consistency at the cost of entanglement with semantic consistency, unlike the synthetic data (right).}
    \end{center}
    
    
\end{figure*}

\section{Related Work}

The seminal work of Gatys's neural style transfer \cite{gatys} introduced the concept of using neural, learning based methods for re-rendering a given content image, such as a photograph, to match the visual artistic style of a second stylistic image, typically an artwork. Other works extended neural style transfer to multiple, and eventually arbitrary styles per model \cite{Karayev2014,ulyanov_multinst,chuanli_multinst,johnson_multinst,wct}.

AvatarNet \cite{avatarnet}, and later SANet \cite{sanet} explore the application of self attention modules in performing style transfer in a feed-forward manner, aligning feature statistics between the content and style images. PAMA \cite{pama} proposes an alternative attention mechanism based on iterative refinement of feature alignment, progressively adjusting content features to match style features in a more spatially consistent manner.
ContraAST \cite{contraAST} uses self attention as per SANet in conjunction with domain-level adversarial losses and contrastive losses to push the stylized images to resemble distributions of real images better - thereby creating more convincingly real looking images regardless of style. CAST \cite{cast} primarily improve this process by including ground truth style images in the contrastive losses.

NeAT \cite{neat} 
further build on the work in ContraAST and CAST, using an expanded version of the attention approach in PAMA, and other robustness and quality improvements. Additionally, they perform stylization as an image editing process rather than an image re-generation process by predicting deltas over a partially corrupted version of the reference content image.

\begin{figure*}
    \begin{center}
    \includegraphics[width=1\linewidth]{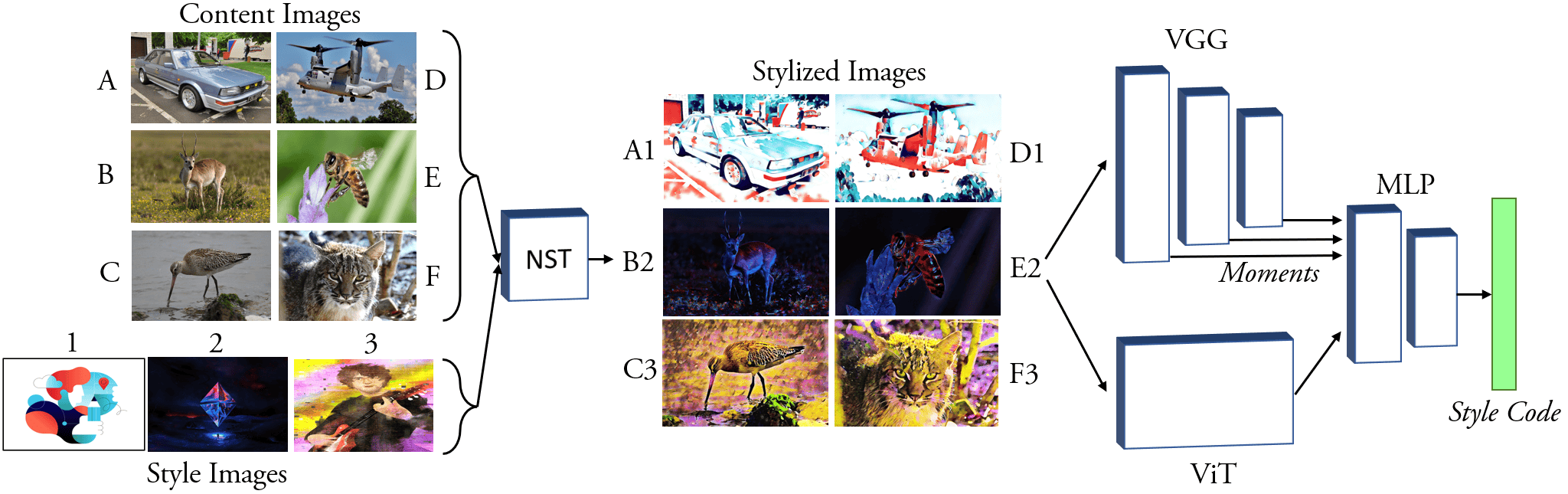}  
    \end{center}
    \caption{Visualization of our NST-driven style representation learning method. We show a training iteration with batch size 6, with 6 content images and 3 style images (in our experiments, we use much larger batch sizes but use 6 here for clarity). The content images are stylized with a pre-trained and frozen Neural Style Transfer method using two copies of the 3 style images. We extract a style embedding using layer-wise global moment statistics and the logits from a more localized vision transformer.
    }
    \label{fig:method}
\end{figure*}

In a similar branch of research, image translation works like MUNIT \cite{munit} and Swapping Autoencoders (SAE) \cite{sae} decompose a pair of images into structural information and global unlocalized latents, which can be mixed during inference to render an image with mixed properties. These works demonstrate how an embedding optimized to capture global information (such as style) can be used in a generative setting.

Using a triplet loss, \cite{Collomosse2017} learn a coarse metric style representation for 7 styles, using the style-labeled subset of the BAM dataset \cite{bam}. ALADIN \cite{aladin} first explored a \textit{fine-grained} style representation using their newly labeled BAM-FG dataset.  
Depending on the chosen similarity strength, this larger dataset contains up to 135k style groups. They design their model for the disentangled representation of content and style by extracting features as global AdaIN statistics from each encoder layer. 
The BAM-FG dataset was curated via crowd annotation to select style-coherent images in existing weakly labeled style coherent groups of images from \textit{Behance.net}. The labeling process removed anomalies and cleaned the coherent style groups such that the remaining images in a group, at several thresholds, was human verified to be style consistent. This helped to drive ALADIN to be state-of-the-art in style representation capabilities, further to the training methodology. 

However, this labeling process needs to be revised. The dataset is indeed style coherent, but the labeling process only improves the style coherency of any given small set of images - it does not help avoid content and style disentanglement. If all the images in a style group have the same content depicted, the BAM-FG cleaning process only helps ensure they are also style consistent. 
But resulting style groups are also consistent in the content information.  

\begin{figure}[H]
    \begin{center}
    \includegraphics[width=0.9\linewidth]{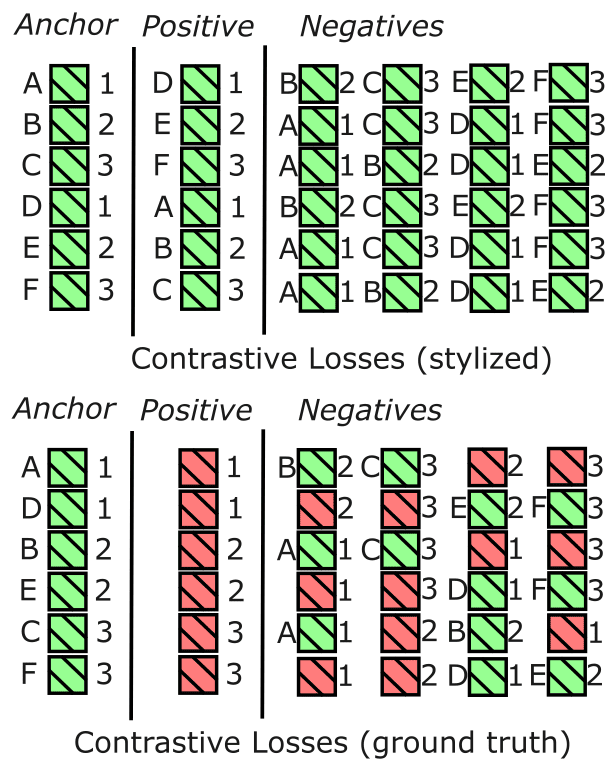}  
    \end{center}
    \caption{A set of contrastive losses are computed for each stylized image in the batch. The positive sample is the other sample in the batch where the same original style image was used as a style reference during stylization. As half the number of style images are selected per batch, there will always be two images with the same style. The negative samples in the contrastive losses are thus the remaining images in the batch, which are stylized with other randomly sampled style images in the batch. Additional sets of contrastive losses compare the stylized images' embeddings to the embeddings of the source style images, as per CAST \cite{cast} and NeAT. Red squares represent ground truth images' embeddings.}
\end{figure}
As artists develop their skills, they likely specialize in specific subsets of subject matter, such as faces or character design. Alternatively, the work they publish can showcase a project they worked on where the subject matter was constrained to some requirements. This effect is visualized in Figure 1 (left), showing a few style groups from the BAM-FG dataset. The images therein are indeed style consistent, but they also share semantic features.

\begin{table*}
  \centering
  \begin{adjustbox}{width=\linewidth}
      \centering
      \small
  
      \begin{tabular}{l|ccc||cc|cc|cc||cc}
        
        Model & \multicolumn{3}{c||}{NST learning signal} & \multicolumn{2}{c|}{NeAT test set}  & \multicolumn{2}{c|}{PAMA test set} & \multicolumn{2}{c||}{SANet test set} & \multicolumn{2}{c}{\textit{Average values}} \\
        & NeAT & PAMA & SANet & mAP & IR-1 & mAP & IR-1 & mAP & IR-1 & mAP & IR-1 \\
        
        \hline
        ALADIN-ViT \cite{stylebabel}  & - & - & - & 16.823 & 0.270 & 9.9964 & 0.0575 & 12.493 & 0.0599 & 13.104 & 0.129 \\
        \hline
        Ours (ViT) & \cmark & &               & 85.306 & 66.308 & 51.012 & 15.303 & 67.525 & 28.423 & 67.948 & 36.678 \\
        Ours (ViT) & & \cmark &               & 69.226 & 23.415 & 62.886 & 20.628 & 51.934 & 6.393 & 61.349 & 16.812 \\
        Ours (ViT) & & & \cmark               & 80.230 & 46.466 & 56.215 & 18.738 & 74.621 & 35.693 & 70.355 & 33.632 \\
        Ours (ViT) & \cmark & \cmark &        & 84.997 & 59.650 & 68.468 & 30.563 & 67.021 & 23.443 & 73.495 & 37.885 \\
        Ours (ViT) & \cmark & & \cmark        & 85.052 & 64.993 & 53.657 & 17.688 & 77.410 & 46.408 & 72.040 & \textbf{43.030} \\
        Ours (ViT) & & \cmark & \cmark        & 77.056 & 36.413 & 64.596 & 23.048 & 67.229 & 22.835 & 69.627 & 27.432 \\
        Ours (ViT) & \cmark & \cmark & \cmark & 83.915 & 58.900 & 67.484 & 29.745 & 74.755 & 34.460 & \textbf{75.385} & 41.035 \\

      \end{tabular}
   \end{adjustbox}

    \caption{
    Style representation learning metrics (IR-topk and mAP) of our model with different NST learning signals. We also compare against ALADIN-ViT \cite{stylebabel}, keeping the same backbone. We measure the representation learning using multiple test sets compiled with different style transfer works from literature: NeAT \cite{neat}, PAMA \cite{pama}, and SANet \cite{sanet}.
    }
    \label{tab:nst_ablations}
\end{table*}

\section{Methodology}

In our work, we set out to create a model to learn disentangled representations of style without being affected by semantics data biases. 
We seek to train a model on data that has high variance in the semantic content depicted but has a consistent style.
As discussed in previous sections, real data with such properties is rare or impractical to create through human artists. Instead, we use the current state-of-the-art neural style transfer methods to create synthetic datasets of stylized images where the style is consistent, but where the content varies depending on our source content images.
Fig. 1 (right) visualizes synthetic stylized data used in our work. Given a style image, we can generate images with the same style but completely random and arbitrary semantic content.

Given a batch of content and style images, we know the synthesized data's ground truth style and content relations. We dynamically use fast, feed-forward NST methods during training to maximize the number of styles we can use without the impractical storage space needed to pre-compute the images. We induce the style learning signal through contrastive losses \cite{simclr}, computed amongst the images generated by the NST method and the reference style image. 
We sample only half the number of style images in a batch to synthesize two images with the same style in each batch, for each style. For each synthetic stylized image, we use the other image stylized with the same style in this batch as the positive and the remaining images in the batch (stylized with the different style images) as negatives. This encourages our embedding to represent the style information shared in the stylized pairs, regardless of the semantic content depicted, which is random. We use standard contrastive losses to drive the learning signal using this self-supervised approach of data labelling, as visualized in Fig \ref{fig:method}.

This approach also benefits from using style images from datasets where style-consistent labeling is not required in a self-supervised manner. We thus use the BBST-4M dataset \cite{neat}, as it has one of the highest diversity of style images - 2 million images in the style subset. The style subset in BBST-4M is also filtered to only contain stylistic (artistic) data, unlike BAM-FG, which includes style groups of non-stylistic images such as photographs. Artistic images are better suited for NST, as these processes are specifically designed to transfer such style.

\subsection{Moments}

Global feature statistics have been demonstrated in literature \cite{adain,wct} to capture global style in an image. Standard statistics used are mean and variance. In moments, these represent the first and second moment, though higher order moments have been used to drive NST through moment matching \cite{moment_matching}, with higher quality. We thus use the first four moments in our work, further extracting skewness and kurtosis from feature statistics in the VGG branch. 

The skewness formula is shown in Eq \ref{eqn:skewness}, calculated via the z scores (Eq \ref{eqn:zscores}), and the kurtosis is shown in Eq \ref{eqn:kurtosis}, where a positive value indicates leptokurtic data distribution, and a negative value indicates a platykurtic distribution - measures of the tails of the data distribution.

\begin{equation}\label{eqn:zscores}
z_{scores} = \frac{X - \mu}{\sigma}
\end{equation}
\begin{equation}\label{eqn:skewness}
m_{3} = \cfrac{\sum {z_{scores}}^3 }{n}
\end{equation}
\begin{equation}\label{eqn:kurtosis}
m_{4} =  \cfrac{ \sum {z_{scores}}^4  }{n} - 3
\end{equation}

We also include highly expressive features extracted from a vision transformer \cite{vit} model to capture more localized features in an image. We concatenate these embeddings and project them into a 1024 dimension style vector, as shown in Figure \ref{fig:method}.

\subsection{Loss}
The loss objective is a standard contrastive loss, shown in Eq \ref{eqn:loss_2}, where $\mathcal{A}$ represents our ALADIN-NST model, $x_s$ and $x_c$ represent style and content images respectively, and $NST$ represents a randomly sampled NST method from the methods used, to stylize $x_s$ and $x_c$ into $\mathcal{S}_{sc}$:

\begin{equation}
    \label{eqn:loss_1}
    \mathcal{S}_{sc} = NST(x_s, x_c)
\end{equation}

\begin{equation}
pos = \mathcal{A}(\mathcal{S}_{sc})_a ^T \mathcal{A}(\mathcal{S}_{sc})_p / \tau
\end{equation}

\begin{equation}
\mathcal{L}:=-\log \left(\frac{\exp \left(pos\right)}{\exp \left( pos \right) + \sum \exp \left(\mathcal{A}(\mathcal{S}_{sc})_a ^T \mathcal{A}(\mathcal{S}_{sc})_n / \tau\right)}\right)
    \label{eqn:loss_2}
\end{equation}

\section{Experiments}

\begin{table*}
  \centering
  \begin{adjustbox}{width=1\linewidth}
      \centering
      \small
  
      \begin{tabular}{l|c||cc|cc|cc||cc}
        
        Model & Dataset & \multicolumn{2}{c|}{NeAT \cite{neat} test set}  & \multicolumn{2}{c|}{PAMA \cite{pama} test set} & \multicolumn{2}{c||}{SANet \cite{sanet} test set} & \multicolumn{2}{c}{\textit{Average values}} \\
        &  & mAP & IR-1 & mAP & IR-1 & mAP & IR-1 & mAP & IR-1 \\
        
        \hline
        ALADIN \cite{aladin}                  & BAM-FG & 59.549 & 8.085 & 38.423 & 1.7025 & 48.712 & 3.560 & 48.895 & 4.449 \\
        $\longrightarrow$ Fused \cite{aladin}  & BAM-FG & 53.941 & 4.485 & 32.686 & 0.550 & 42.592 & 2.395 & 43.073 & 2.477 \\
        ALADIN-ViT \cite{stylebabel}               & BAM-FG & 16.823 & 0.270 & 9.996 & 0.058 & 12.493 & 0.060 & 13.104 & 0.129 \\
        SAE \cite{sae}                      & BAM-FG & 51.600 & 16.100 & 28.500 & 4.000 & 28.814 & 4.643 & 36.305 & 8.248 \\
        \hline

        Ours                     & BAM-FG & 85.955 & 58.108 & 67.699 & 24.967 & 74.355 & 27.154 & 76.003 & 36.743 \\
        Ours                     & BBST-4M & 90.965 & 69.523 & 80.861 & 42.803 & 84.953 & 45.258 & \textbf{85.593} & \textbf{52.528} \\
        
      \end{tabular}
   \end{adjustbox}

    \caption{
    Style representation strength, compared to baseline methods. Higher values are better.
    }
    \label{tab:style_baselines}
\end{table*}

\begin{table*}
  \centering
  \begin{adjustbox}{width=1\linewidth}
      \centering
      \small
  
      \begin{tabular}{l|c||cc|cc|cc||cc}
        
        Model & Dataset & \multicolumn{2}{c|}{NeAT \cite{neat} test set}  & \multicolumn{2}{c|}{PAMA \cite{pama} test set} & \multicolumn{2}{c||}{SANet \cite{sanet} test set} & \multicolumn{2}{c}{\textit{Average values}} \\
        &  & mAP & IR-1 & mAP & IR-1 & mAP & IR-1 & mAP & IR-1 \\
        
        \hline
        ALADIN \cite{aladin}                   & BAM-FG & 5.547 & 0 & 8.523 & 0 & 4.642 & 0 & 6.237 & \textbf{0} \\
        $\longrightarrow$ Fused \cite{aladin}  & BAM-FG & 11.008 & 0 & 19.239 & 0.013 & 8.920 & 0 & 13.056 & 0.004 \\
        ALADIN-ViT \cite{stylebabel} & BAM-FG & 15.058 & 0.023 & 10.081 & 0.028 & 8.097 & 0.003 & 11.079 & 0.018 \\
        SAE \cite{sae}                      & BAM-FG & 2.198 & 0.003 & 3.815 & 0.005 & 2.758 & 0 & 2.924 & 0.003 \\
        \hline
        Ours                     & BAM-FG & 1.523 & 0 & 1.575 & 0 & 1.630 & 0 & 1.576 & \textbf{0} \\
        Ours                     & BBST-4M & 1.491 & 0 & 1.427 & 0 & 1.652 & 0 & \textbf{1.523} & \textbf{0} \\

      \end{tabular}
   \end{adjustbox}

    \caption{
    Comparisons with the same baselines as Table \ref{tab:style_baselines}, measuring similarity between content images. In this table, a higher value is worse, as it represents higher content entanglement. 
    }
    \label{tab:content_baselines}
\end{table*}

Due to the cross-NST approach in our style representation learning signal, the 2 million images in each of the content and style splits of BBST-4M lead to a synthetic dataset of an effective 4 \textit{trillion} images. This creates a practically limitless combination of style and content during training.

\subsection{Data}

To evaluate how well the model represents specifically disentangled style information, we need to consider test data that is also wholly disentangled. 
We also apply our synthetic NST dataset creation methodology for the test set, ensuring no overlap with training data.
Using 400 new style images from Behance, and 100 new content images from Flickr, we extend BBST-4M with synthetic stylized images. We create 40k images, stylizing each content image with each style.

We use NeAT, PAMA, and SANet variants of this test set to evaluate the generalization of style representation independent of any systematic signatures specific to any NST method - visible or otherwise.
We select these three NST methods given their fast and leading stylization qualities in literature. We manually selected the source style images to ensure a high variety of styles and no duplicates or styles too similar by manually inspecting style-based image retrieval for each style image as a query over the remaining test set style source corpus.

We'd like to stress that although we are evaluating the disentanglement capabilities of our technique on synthetic data, we show through our other experiments that our representation carries over its strengths to real data, also. The synthetic data is only used for evaluating disentanglement properties, as real disentangled data is not available.

\subsection{Metrics}
We build our evaluation pipeline around image retrieval using these synthetic test sets. The primary metric we measure is mean average precision (mAP). We calculate the mAP by considering the other 99 content images stylized with the same style as positives and those stylized with the different style images as negatives. For each image in the test set, we re-arrange the remaining test set images, sorted by similarity in the style embedding space to this query image.
We also use the Instance Retrieval (IR) \cite{aladin} metric from ALADIN, for which we measure the \textit{top-k} accuracy of retrieving the source style image from the corpus. We remove the other stylized images of the same style from the corpus to only leave the query and source images sharing the same style for a given search.

\begin{figure*}
  \centering
  \small
      \centering
      \begin{adjustbox}{width=1\textwidth}
          \begin{tabular}{p{0.21\linewidth}p{0.20\linewidth}p{0.20\linewidth}p{0.20\linewidth}p{0.20\linewidth} }

            \vspace{-0.2cm} \raisebox{-\totalheight}{\includegraphics[width=0.21\textwidth]{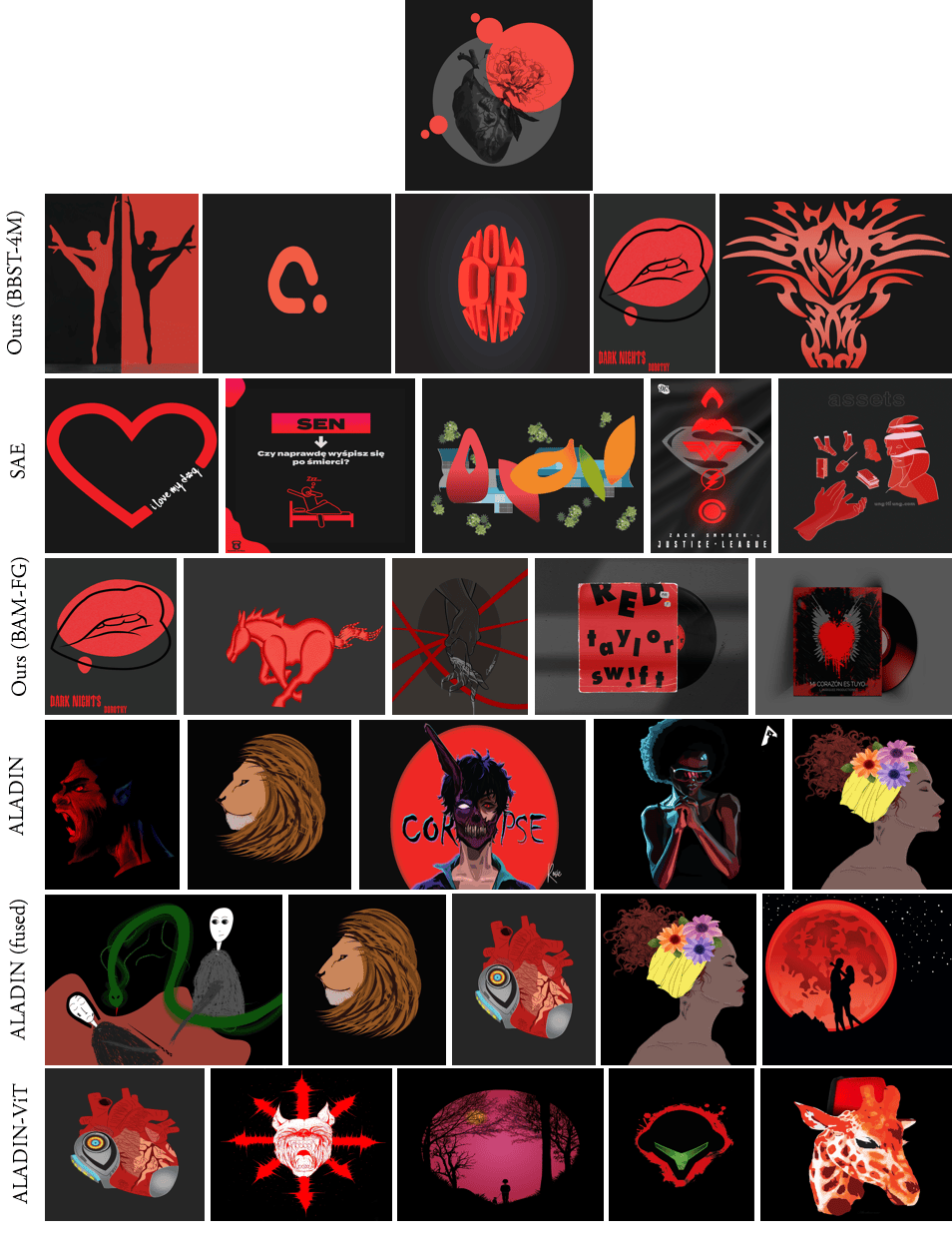}} & 
            \vspace{-0.2cm} \raisebox{-\totalheight}{\includegraphics[width=0.21\textwidth]{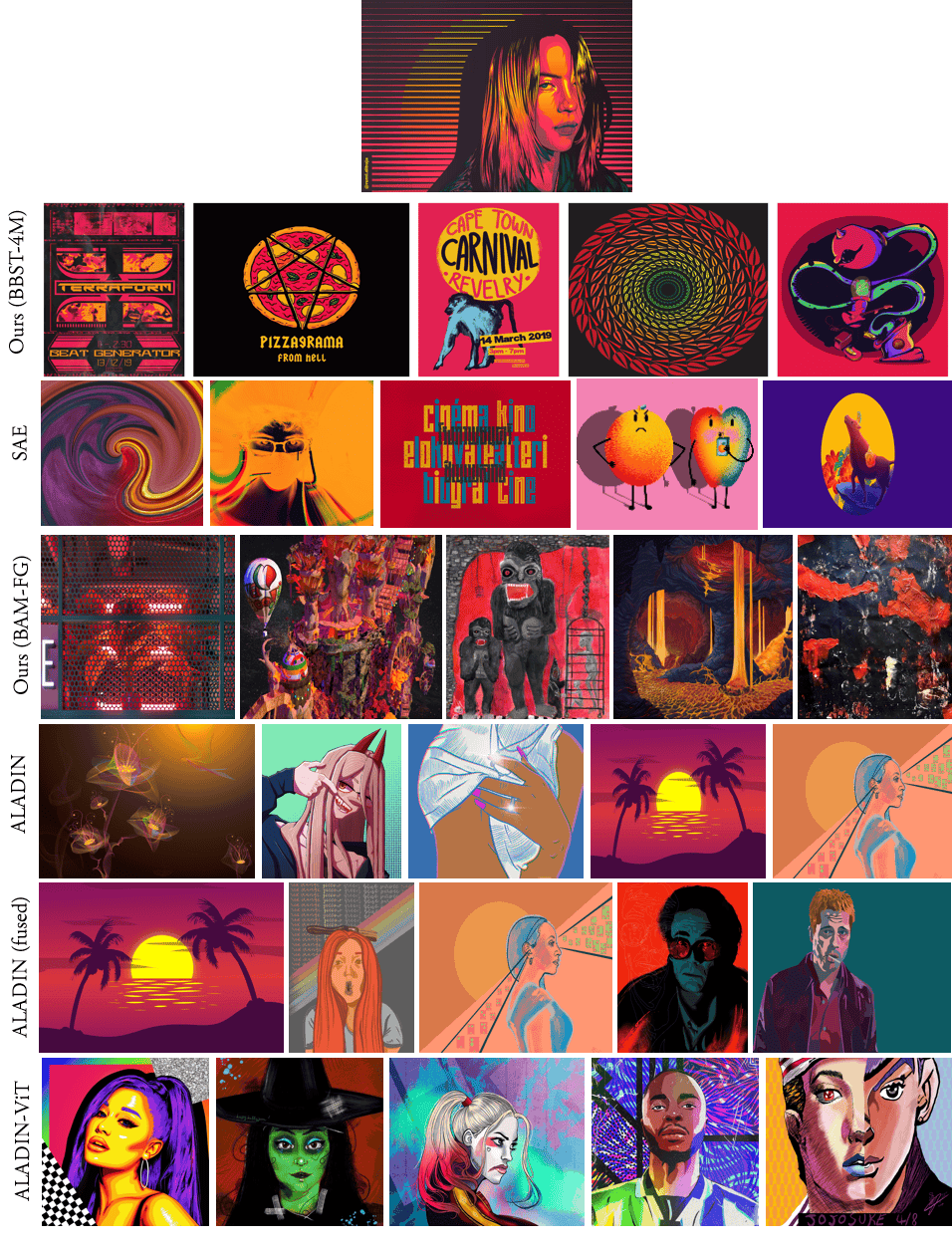}} \\ 

            \vspace{-0.2cm} \raisebox{-\totalheight}{\includegraphics[width=0.21\textwidth]{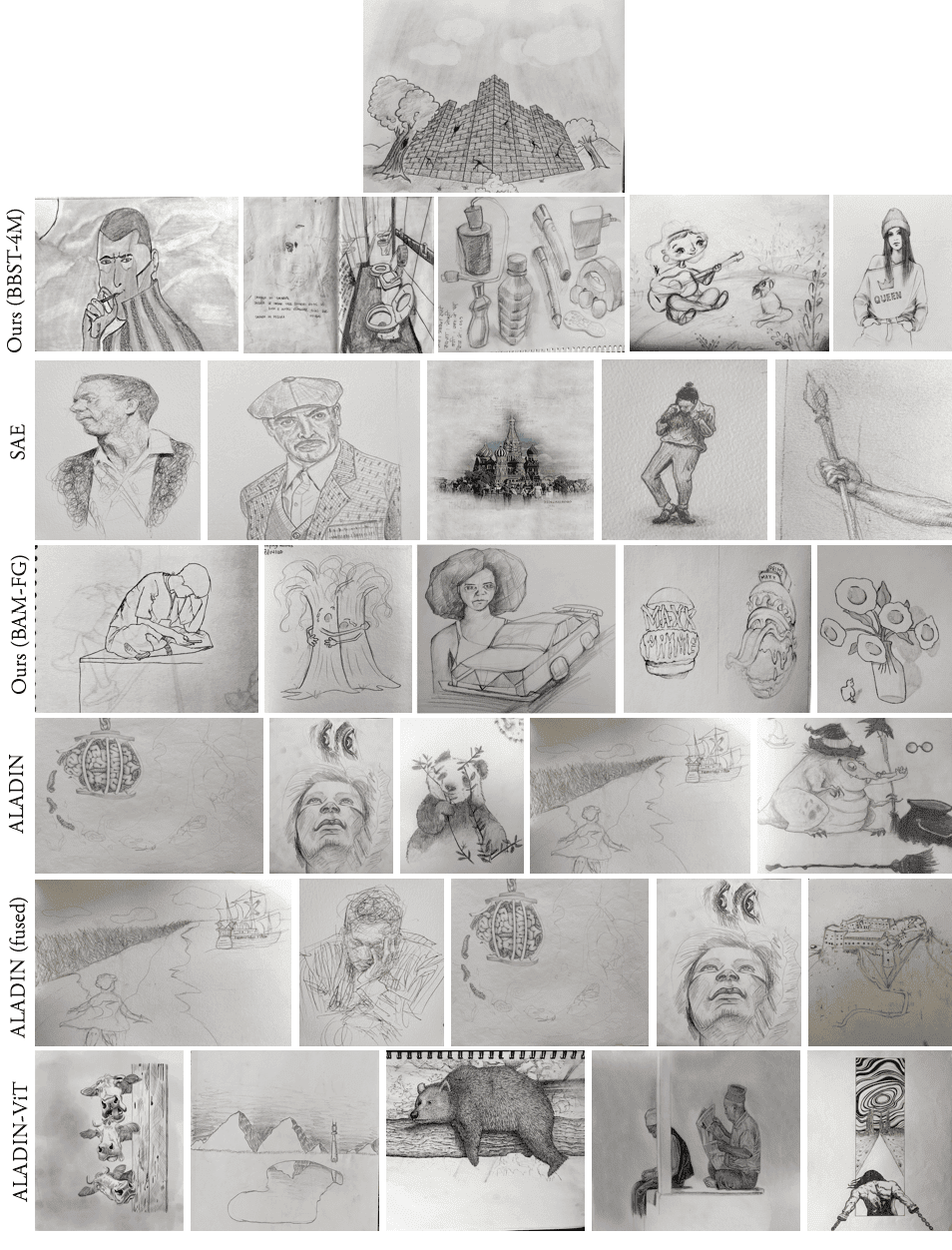}} & 
            \vspace{-0.2cm} \raisebox{-\totalheight}{\includegraphics[width=0.21\textwidth]{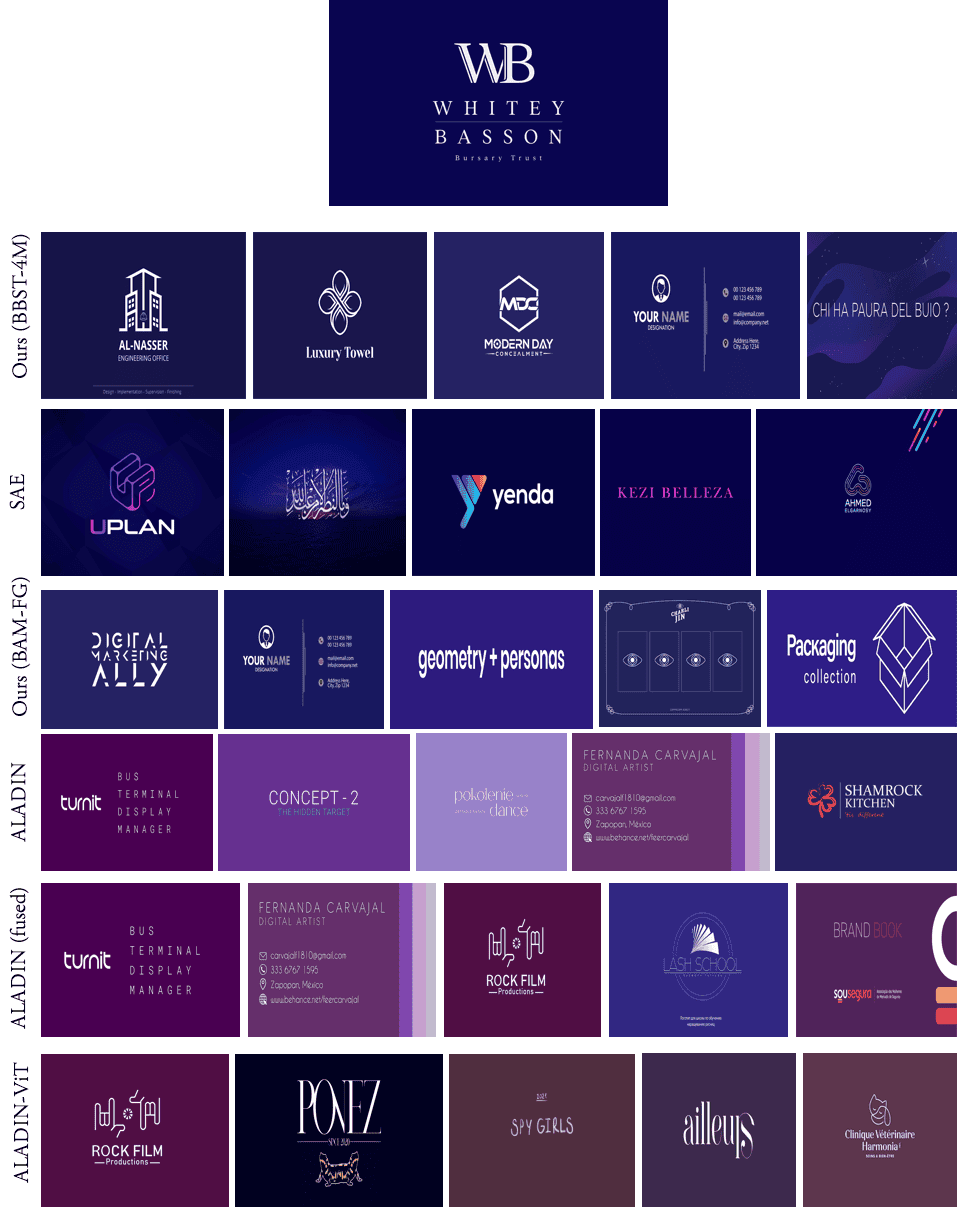}} \\

          \end{tabular}
      \end{adjustbox}
    \caption{Style-based image retrieval comparison between our method variants and previous literature.}
    \label{fig:img_ret}
\end{figure*}
\begin{figure*}
  \centering
  \squeezeup
  \small
      \centering
      \begin{adjustbox}{width=0.95\textwidth}
          \begin{tabular}{p{0.12\linewidth}p{0.21\linewidth}p{0.20\linewidth}p{0.20\linewidth}p{0.20\linewidth}p{0.20\linewidth} }

            &
            \vspace{-0.2cm} \raisebox{-\totalheight}{\includegraphics[width=0.20\textwidth]{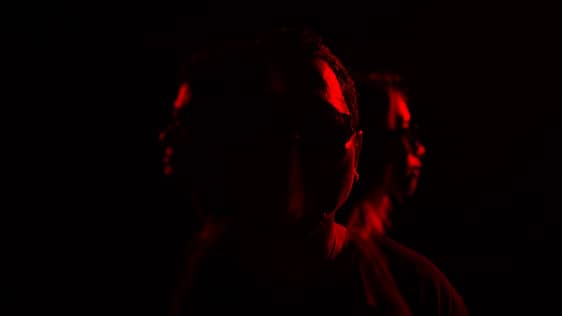}} & 
            \vspace{-0.2cm} \raisebox{-\totalheight}{\includegraphics[width=0.20\textwidth]{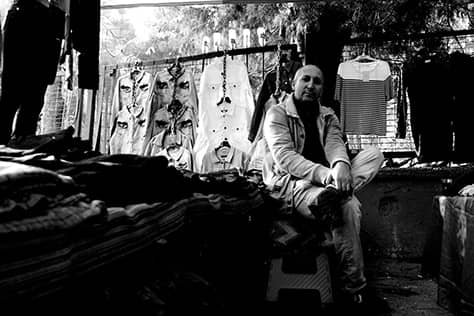}} & 
            \vspace{-0.2cm} \raisebox{-\totalheight}{\includegraphics[width=0.20\textwidth]{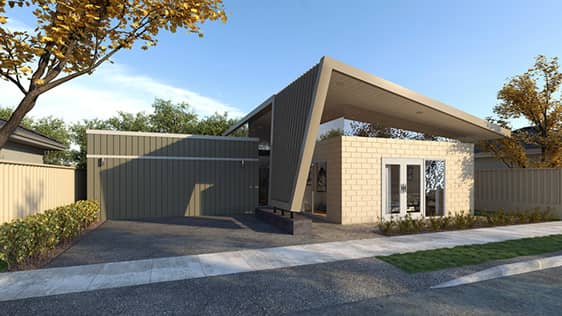}} & 
            \vspace{-0.2cm} \raisebox{-\totalheight}{\includegraphics[width=0.20\textwidth]{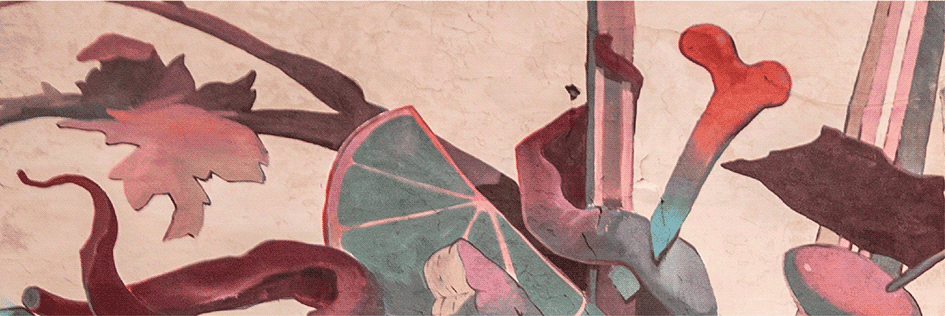}} & 
            \vspace{-0.2cm} \raisebox{-\totalheight}{\includegraphics[width=0.20\textwidth]{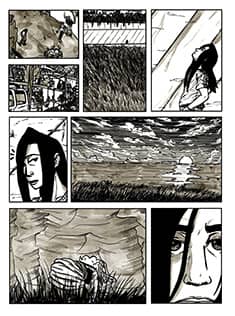}} \\

            ALADIN-ViT &
            neon-like, warm toned lighting, light trail, strong outline, all cap &
            narrative driven, manga, Japanese, cartoon drawing, graphic art &
            regulated layout, posh, promoting, triangle composition, branding package & 
            mixed media, layered composition, watercolor painting, handmade artwork painting, illustrative &
            color splash, marker drawing, expressionist, word sound effects, subtle colors  
             \\
            
            \hline
            Ours (fused) &
            dark image, chiaroscuro, black and red, red, dark picture & 
            documentary, vigorous, past, documentary shot, dark contrast & 
            trustful, housing, posh, housing architecture render, solarpunk & 
            colorful drawing, abstract art, abstract artwork, nonobjective, cubism & 
            storyboarding, interpretive, panel, story, storyboarding \\
             
            \hline
            Ours  &
            flame, spark, dark vibe, explosion, dark space & 
            contrasted, high contrast, suburban, documentary shot, documentary & 
            glass, trustful, architectural landscape, pentagonal, pointy & 
            soft, soft color, soft and bright color, feminine, soft colors & 
            comic book art, black and white art, comic art, comic, doodle art

            \\
            &
            \vspace{-0.2cm} \raisebox{-\totalheight}{\includegraphics[width=0.20\textwidth]{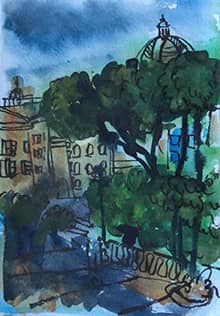}} & 
            \vspace{-0.2cm} \raisebox{-\totalheight}{\includegraphics[width=0.20\textwidth]{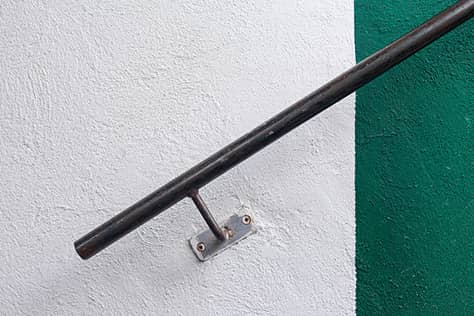}} & 
            \vspace{-0.2cm} \raisebox{-\totalheight}{\includegraphics[width=0.20\textwidth]{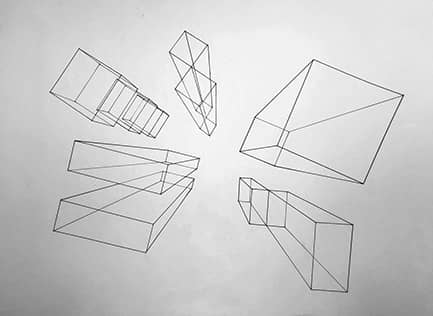}} & 
            \vspace{-0.2cm} \raisebox{-\totalheight}{\includegraphics[width=0.20\textwidth]{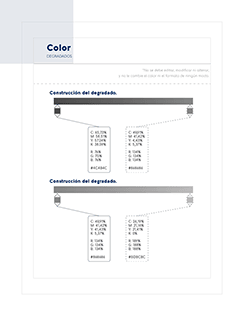}} & 
            \vspace{-0.2cm} \raisebox{-\totalheight}{\includegraphics[width=0.20\textwidth]{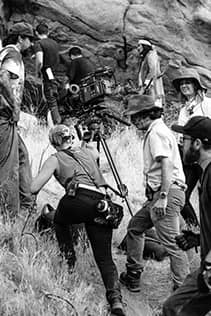}} \\

            ALADIN-ViT &
            cold hue, product-focused, product description, blue-based, digital publication &
            had material, watery, bird eye view, nobody & 
            measuring, technical sketch, design sketch, sketch, typography element & 
            user input, 3d building plan, internet, gathering information, datum collecting  & 
            expressive, irregular angle, clean stroke, blue glow, copy  
             \\
            
            \hline
            Ours (fused) &
            mystical, fantasy concept art, fantasy art, aqua, fantasy painting &
            reflective, commercial shot, rough texture, geometric shape, geometric line & 
            sketch scamp, scamp, sketch work, sketch, sketched line & 
            uxui design, design interface, user interface, ui instructional design, interface & 
            poverty, struggle, slum, evocative, documented \\ 
             
            \hline
            Ours  &
            layered composition, aqua, mystical, fantasy painting, digital print &
            embossed, small shape, cup, stone image, light grey & 
            idea, blended, pen and pencil, sketch, quick & 
            page layout, image of article, blocky layout, conceptual layout, stationery design & 
            high-contrast, retouched, noir, poverty, slum  
                         
            \\
            &
            \vspace{-0.2cm} \raisebox{-\totalheight}{\includegraphics[width=0.20\textwidth]{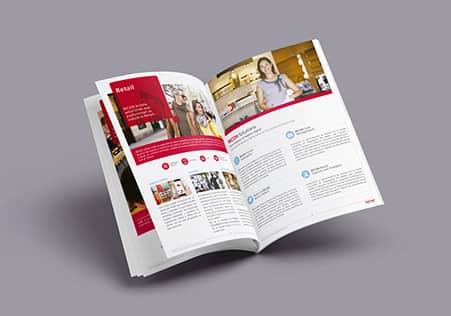}} & 
            \vspace{-0.2cm} \raisebox{-\totalheight}{\includegraphics[width=0.20\textwidth]{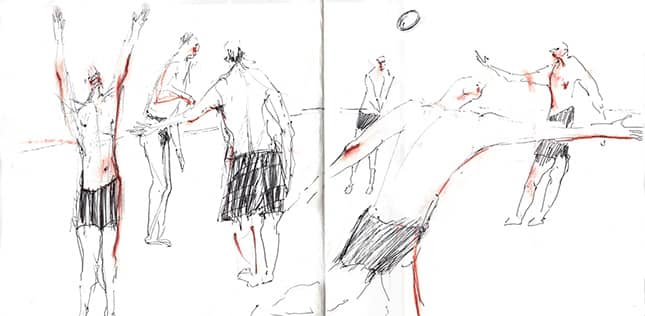}} & 
            \vspace{-0.2cm} \raisebox{-\totalheight}{\includegraphics[width=0.20\textwidth]{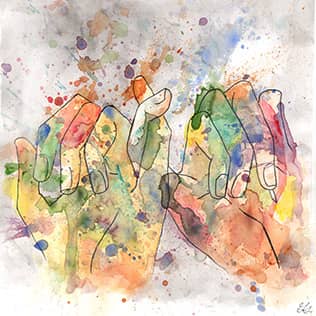}} & 
            \vspace{-0.2cm} \raisebox{-\totalheight}{\includegraphics[width=0.20\textwidth]{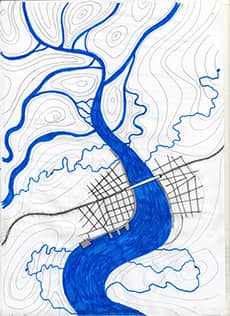}} & 
            \vspace{-0.2cm} \raisebox{-\totalheight}{\includegraphics[width=0.20\textwidth]{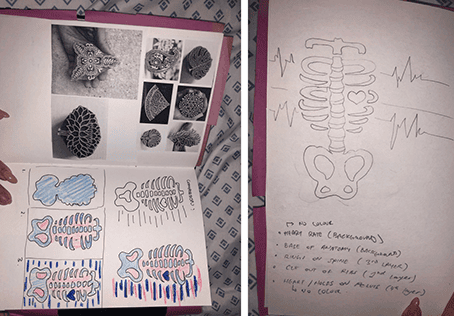}} \\

            ALADIN-ViT &
            cool hue, mockup, magazine booklet layout, blue themed, graphic layout &
            hand drawn, changing proportion, sketchy, saddening, line drawing & 
            colorful, bright, bright bold, flow, color-heavy & 
            curved line, colorful drawing, mark making, blue ink drawing, thin stroke & 
            fading, sans serif and serif, red highlighting, thin letter, typeface  
              \\
            
            \hline
            Ours (fused) &
            editorial design, editorial, editorial mockup, editorial work, editorial mockup &
            line, ink work, ink, ink drawing, outline & 
            pastel, pastoral, black tea painting, art appreciation, colorful drawing & 
            abstract line, mark-making, line, fine, fluid line & 
            spontaneous, beginner, sketchbook, paper pattern, light pencil work \\ 
             
            \hline
            Ours  &
            editorial, editorial design, readable, editorial mockup, professional style &
            animation sketch, fine-line drawing, figurative drawing, line drawing illustration, sketch of cartoon character & 
            pastel, art therapy, fine art, traditional illustration, child illustration & 
            fluid, blue,cut paper, plump, material sample & 
            printmaking, delicate type, various printed paper work, lino, handmade  
                        
            \\

          \end{tabular}
      \end{adjustbox}
      \squeezeup
      \squeezeup
    \caption{Please zoom for more image detail. Zero-shot automatic style tagging comparison, between ALADIN-ViT, our model, and our fused variant, joining our disentangled embeddings with ALADIN-ViT. We show the top 5 tags for each image.}
    \label{fig:tags_viz}
    \squeezeup
    \squeezeup
\end{figure*}

\subsection{Ablations}

Table \ref{tab:nst_ablations} contains ablations where we experiment with the NST methods used for driving the style learning signal during training. Using more than one method is essential for generalizing the style representation. Using only one NST method risks modeling specific artifacts of that method. Future work could incorporate additional new NST methodologies as the field advances. Our final model uses all 3 methods: NeAT, PAMA, and SANet. We explore these ablations using only a ViT backbone, such that we can also draw fair comparisons to literature using the same architecture, ALADIN-ViT.

\subsection{Baselines}

We compare our model against baselines in Table \ref{tab:style_baselines}. 
Demonstrating also how the BAM-FG dataset needs to be revised as a style-only dataset. 
We train Swapping Autoencoders (SAE) \cite{sae}, and our model with BAM-FG to compare the data fairly.

The accuracy rankings according to the literature are flipped in our measurements as we test with purely disentangled labels. ALADIN scores highest despite being the first relevant model because its features are extracted globally. Their fusion includes ResNet embeddings, which introduce semantic entanglement as a by-product of the higher BAM-FG style scores.
ALADIN-ViT scores are even lower on our disentangled test set due to a lack of explicitly global features, therefore more intensely focusing on localized and, thus, typically more semantic information.

In Table \ref{tab:content_baselines}, we repeat our evaluation found in Table \ref{tab:style_baselines}, but instead of measuring the style retrieval in our test sets, we measure \textit{content} retrieval. Like the style evaluation, we compute mAP by using a query image and evaluating retrieval of the corpus concerning all the other 399 images of the same content, but stylized with different style images. In other words, measuring semantics-based image retrieval, irrespective of style. 

For IR-\textit{k}, we filter out the other stylized versions of the content image stylized in the query and measure the retrieval of the original un-stylized content image. We run this set of evaluations to measure how strongly the style embeddings capture content/semantic information. Supporting our previous explanations, ALADIN's fused and ViT variants each capture more semantic information. Our work improves upon this, as our style embeddings perform much more poorly at retrieving images with the same content.

\subsection{Training details}
We train ALADIN-NST with the NeAT, PAMA, and SANet NST methods for roughly 3 days on a single A100 GPU until convergence. We stylized images for training using 512px resolution, which we downsample to 256x256 for the VGG branch and 224x224 as needed for ViT-B\_16, using the same ViT as ALADIN. We disable the prior blurring in NeAT for speed. We use the Adam optimizer, and a target batch size of 1024 via logit accumulation. We decay the learning rate by 0.999875 every 100 iterations.

\subsection{Style-based image retrieval}

In Figure \ref{fig:img_ret}, we visualize a comparison of our method to baseline methods in the literature for style-based image retrieval. We show our approach trained on both BBST-4M and BAM-FG. We perform the retrieval over a corpus of 500k images from BBST-4M. 

Our results are comparable regarding visual features, improving slightly on the color consistency of retrieved results (bottom right). However, there is less semantic consistency between our model's query and outcomes, especially compared to the previous ALADIN models. In the top left of Figure \ref{fig:img_ret}, the \textit{heart} in the query image retrieves some other heart-related imagery in baselines. In the top right, baseline recovered results contain faces and character designs, also present in the query.

\section{Multimodal vision-language learning}

We use style embeddings from our proposed model to learn a joint multimodal representation between style and language. We replicate the work in StyleBabel \cite{stylebabel}, where style tags attributed to images can be used as labels for this task. We replace their ALADIN-ViT vision backbone with ours, and we similarly train an MLP, joining these style embeddings to text embeddings extracted using CLIP \cite{clip} through contrastive learning. We aim to measure how our new style embeddings can be used in this multimodal setting.

We measure a WordNet score of 0.329, which beats their baseline CLIP WordNet score of 0.215, but does not beat the ALADIN-ViT WordNet score of 0.352. This may be due to the inherent content/style entanglement of the style tags in StyleBabel, which itself is not strictly disentangled. There exist several tags in StyleBabel, such as \textit{t-shirt design}, \textit{interior design}, \textit{fashion photography}, which do describe the style, but in a context that is also grounded in semantics. By explicitly not encoding semantic information in our style embeddings, such tags are more difficult to retrieve.  

Inspired by the fusing \cite{aladin} of the complementary ALADIN and ResNet \cite{resnet} embeddings, we explore a fusion of our disentangled style embeddings with ALADIN-ViT embeddings, which contain some semantic information. We extract and concatenate embeddings from both models and use this dual-model embedding as a style embedding for learning the joint vision+language multimodal embedding against CLIP. We achieve a state-of-the-art WordNet score of \textbf{0.415} on StyleBabel tags. We use the same test split for our measurements. In Fig. \ref{fig:tags_viz}, we visualize automatic zero-shot style tagging with ALADIN-ViT (baseline), our model, and our model fused with ALADIN-ViT over images from the StyleBabel test set.

\section{Conclusions}

We explore a novel learning methodology for artistic style, achieving stronger disentanglement. We demonstrate the value of this by further achieving state-of-the-art multimodal vision+language learning on StyleBabel tags.

Our approach relies on NST as a strong driver of style consistency, thus limiting us to the capabilities of such automated stylization methods. 
However, as NST methods continue to improve, so can our method in how well it can capture style. The better the artistic stylization process can be modeled by NST models, the better our technique can be trained to capture that style, by simply including the technique in our pipeline. 

For practicality, we can only rely on fast, feed-forward approaches. Optimization and diffusion based techniques are too slow to dynamically synthesize training data during the training loop unless this data is synthesized ahead of time, trading off variety of artistic styles and high storage costs.

Further work could explore scaling the Vision Transformer branch of the model, as well as exploring variants with a more global context.

{\small
\bibliographystyle{ieee_fullname}
\bibliography{main}
}

\end{document}